\def\year{2022}\relax
\pdfoutput=1
\documentclass[letterpaper]{article} 
\usepackage{aaai22}  
\usepackage{times}  
\usepackage{helvet}  
\usepackage{courier}  
\usepackage[hyphens]{url}  
\usepackage{graphicx} 
\usepackage{natbib}  
\usepackage{caption} 
\DeclareCaptionStyle{ruled}{labelfont=normalfont,labelsep=colon,strut=off} 
\usepackage{algorithm}
\usepackage{algorithmic}

\usepackage{amsmath}
\usepackage{amsfonts}
\usepackage{booktabs}

\usepackage{newfloat}
\usepackage{listings}
\floatstyle{ruled}
\newfloat{listing}{tb}{lst}{}
\floatname{listing}{Listing}
 \usepackage{color} 
 \usepackage{soul}

\title{Sequence-to-Action: Grammatical Error Correction \\
with Action Guided Sequence Generation}
\author {
    Jiquan Li\textsuperscript{\rm 1},
    Junliang Guo\textsuperscript{\rm 1},
    Yongxin Zhu\textsuperscript{\rm 1},
    Xin Sheng\textsuperscript{\rm 1},
    Deqiang Jiang\textsuperscript{\rm 2},
    Bo Ren\textsuperscript{\rm 2},
    Linli Xu\textsuperscript{\rm 1}
}
\affiliations {
    \textsuperscript{\rm 1}Anhui Province Key Laboratory of Big Data Analysis and Application, \\School of Computer Science and Technology, University of Science and Technology of China\\
    \textsuperscript{\rm 2}Tencent YouTu Lab\\
    \{lijiquan, guojunll, zyx2016, xins\}@mail.ustc.edu.cn, \{dqiangjiang, timren\}@tencent.com, linlixu@ustc.edu.cn
}

\usepackage{bibentry}

\begin{document}

\maketitle

\begin{abstract}
The task of Grammatical Error Correction (GEC) has received remarkable attention with wide applications in Natural Language Processing (NLP) in recent years.
While one of the key principles of GEC is to keep the correct parts unchanged and avoid over-correction, previous sequence-to-sequence (seq2seq) models generate results from scratch, 
which are not guaranteed to follow the original sentence structure and may suffer from the over-correction problem.
In the meantime, the recently proposed sequence tagging models can overcome the over-correction problem by only generating edit operations, but are conditioned on human designed language-specific tagging labels.
In this paper, we combine the pros and alleviate the cons of both models by proposing a novel Sequence-to-Action~(S2A) module. The S2A module jointly takes the source and target sentences as input, and is able to automatically generate a token-level action sequence before predicting each token, where each action is generated from three choices named SKIP, COPY and GENerate.
Then the actions are fused with the basic seq2seq framework to provide final predictions. We conduct experiments on the benchmark datasets of both English and Chinese GEC tasks.
Our model consistently outperforms the seq2seq baselines, while being able to significantly alleviate the over-correction problem as well as holding better generality and diversity
in the generation results compared to the sequence tagging models.
\end{abstract}

\section{Introduction}

Grammatical Error Correction (GEC), which aims at automatically correcting various kinds of errors in the given text,  has received increasing attention in recent years, with wide applications in natural language processing such as post-processing the results of Automatic Speech Recognition~(ASR) and Neural Machine Translation~(NMT) models~\citep{9053126}, as well as providing language assistance to non-native speakers.
In the task of GEC, the primary goals are two-fold: 1) identifying as many errors as possible and successfully correcting them; 2) keeping the original correct words unchanged without bringing in new errors.

Intuitively, GEC can be considered as a machine translation task by taking the incorrect text as the source language and the corrected text as the target language. 
In recent years,  NMT-based approaches~\citep{DBLP:journals/corr/BahdanauCB14} with sequence-to-sequence (seq2seq) architectures have become the preferred solution 
for the GEC task, where the original sentences with various kinds of grammatical errors are taken as the source input while the correct sentences are taken as the target supervision.  
Specifically, the Transformer model~\citep{DBLP:journals/corr/VaswaniSPUJGKP17,bryant-etal-2019-bea} has been a predominant choice for NMT-based methods.

However, there are some issues with the seq2seq methods for the GEC task. In general, as a result of generating target sentences from scratch, the repetition and omission of words frequently occur in the seq2seq generation process as studied by previous works~\citep{tu-etal-2016-modeling}. More importantly, there is no guarantee that the generated sentences can keep all the original correct words while maintaining the semantic structures. Experiments show that such problems occur more frequently when the original sentences are too long or contain low-frequency/out-of-vocabulary words.
Consequentially, the seq2seq models may suffer from these potential risks in practice which conflict with the second primary goal of the GEC task.
Recently, sequence tagging methods \citep{malmi2019encode,awasthi2019parallel,omelianchuk2020gector,stahlberg2020seq2edits} consider GEC as a text editing task by detecting and applying edits to the original sentences, therefore bypassing the above mentioned problems of seq2seq models. Nevertheless, the edits are usually constrained by 
human designed or 
automatically generated lexical rules~\citep{omelianchuk2020gector,stahlberg2020seq2edits} and vocabularies~\citep{awasthi2019parallel,malmi2019encode}, which limits the generality and transferability of these methods.
Moreover, when it comes to corrections which need longer insertions, most of these sequence tagging methods rely on iterative corrections, which 
can reduce the fluency.

To tackle these problems, in this paper, we propose a Sequence-to-Action~(S2A) model which is able to automatically edit the erroneous tokens in the original sentences without relying on human knowledge, while keeping as many correct tokens as possible.
Specifically, we simply introduce three atomic actions named \texttt{SKIP}, \texttt{COPY} and \texttt{GEN}erate to guide the model when generating the corrected outputs. 
By integrating the erroneous source sentence and the golden target sentence, we construct a tailored input format for our model, and for each token, the model learns to skip it if the current input token is an erroneous one, or copy it if the token is an originally correct token, or generate it if the token is a target token.

In this way, the proposed model 
provides the following benefits. 
1) As a large proportion of tokens are correct in the source sentences, the \texttt{COPY} action will appear more frequently than the other actions.
Therefore, by taking the source sentence as an input to the S2A module, our model is more inclined to copy the current input token, and therefore alleviates the over-correction problem of seq2seq models.
2) Comparing to sequence tagging models
which generate edits based on the human-designed rules or vocabularies, our model can realize more flexible generations such as long insertions. 
More importantly, our model is language-independent with good generality.

We evaluate the S2A model on both English and Chinese GEC tasks. 
On the English GEC task, the proposed method achieves $52.5$ in $F_{0.5}$ score, producing 
an improvement of $2.1$ over the baseline models. Meanwhile, it achieves $38.06$ in $F_{0.5}$ score on the Chinese GEC task, with improvement of $1.09$ over the current state-of-the-art baseline method MaskGEC~\citep{DBLP:conf/aaai/ZhaoW20}.

\section{Related Work}

\subsection{GEC by Seq2seq Generation}

A basic seq2seq model for GEC consists of an encoder and a decoder, where the source sentence is first encoded into the corresponding hidden states by the encoder, and each target token is then generated by the decoder conditioned on the hidden states and 
the previously generated tokens.
Given a pair of training samples $(x,y)$, the objective function of the seq2seq based GEC model can be written as:
\begin{equation}
\label{equ:s2s-obj}
    \mathcal{L}_{\textrm{s2s}}(y|x;\Theta) = -\sum_{i=1}^n \log p(y_i | y_{<i}, x; \theta_{\textrm{s2s}}),
\end{equation}
where $\theta_{\textrm{s2s}}$ indicates the model parameters.
In general, the input and output sequences may overlap significantly for the GEC task, while a seq2seq GEC model generates the target sequences from scratch given that no source token is directly visible to the decoder, which may lead to over-correction or generation errors.

In the past few years, several methods have been proposed to improve the performance of seq2seq GEC models. \citet{kaneko-etal-2020-encoder} incorporate a pre-trained masked language model such as BERT~\citep{devlin-etal-2019-bert} into a seq2seq model, 
which is then fine-tuned on the GEC dataset.
A copy-augmented architecture is proposed in~\citep{zhao-etal-2019-improving} for the GEC task by copying the unchanged words from the original sentence to the target sentence. Recent advances in seq2seq GEC models mainly focus on constructing additional synthetic data for pre-training~\citep{grundkiewicz-etal-2019-neural, xie-etal-2018-noising, ge-etal-2018-fluency, kiyono-etal-2019-empirical, zhou-etal-2020-improving-grammatical} by directly adding noise to normal sentences, through back-translation, or by using poor translation models.
Nevertheless, even with the above improvements, the seq2seq GEC models still suffer from generating results from scratch, which inevitably leads to over-correction and generation errors.

\subsection{GEC by Generating Edits}

Another line of research takes a different perspective by treating the GEC task as a text editing problem and proposes sequence tagging models for GEC.
In general, these models predict an edit tag sequence $t = (t_1, t_2, ..., t_m)$ based on the source sentence $x$ by estimating $p(t | x) = \prod_{i=1}^m{p(t_i | x)}$. In the edit tag sequence, each tag is selected from a pre-defined set and assigned to a source token,
representing an edition to this token.

Specifically, LaserTagger~\citep{malmi2019encode}
predicts editions between \textit{keeping}, \textit{deleting} or \textit{adding} a new token/phrase from a handcrafted vocabulary.
PIE~\citep{awasthi2019parallel} iteratively predicts token-level editions for a fixed number of iterations in an non-autoregressive way. 
GECToR~\citep{omelianchuk2020gector} further improves the model by designing more fine-grained editions w.r.t the lexical rules of English.
Seq2edits~\citep{stahlberg2020seq2edits} generates span-level (instead of token-level) tags to generate more compact editions.
In the sequence tagging models mentioned above, the operations of generating new tokens 
are restricted either by  human-designed vocabularies or language-dependent lexical rules, 
limiting their generality. For example, GECToR outperforms baseline models on English GEC datasets, but its performance severely degenerates on 
the Chinese GEC task, as shown in Table~\ref{tab:nlpcc_res}.
There also exist methods that integrate seq2seq models with sequence tagging methods into a pipeline system to improve the performance or efficiency~\citep{chen-etal-2020-improving-efficiency,hinson-etal-2020-heterogeneous}, which cannot be optimized end-to-end.

Different to the works discussed above, the proposed S2A model alleviates the over-correction/omission problem of seq2seq models by taking the original sentence as input and  keeping as many correct tokens as possible. Meanwhile, by generating corrections with a set of actions including  \texttt{SKIP}, \texttt{COPY} and \texttt{GEN}erate which do not rely on any human-designed rules or vocabularies, S2A is language-independent and more flexible compared to text editing models.

\section{Methodology}

In this section, we elaborate on the proposed 
model for GEC. We start with the problem definition, followed with the motivation and the model architecture. 
We then proceed to demonstrate how the model is designed.

\noindent \textbf{Problem Definition}~~~
Given a parallel training dataset $(\mathcal{X},\mathcal{Y})$ which consists of pairs of the original erroneous source and the golden target sequences $(x,y) \in (\mathcal{X}, \mathcal{Y})$, where $x = (x_1, x_2, ..., x_m)$ and $y = (y_1, y_2, ..., y_n)$, we aim at correcting $x$ to $y$ through the proposed model by optimizing $p(y|x)$.

\subsection{Model Architecture}

\begin{figure}
    \centering
    \includegraphics[width=1\linewidth]{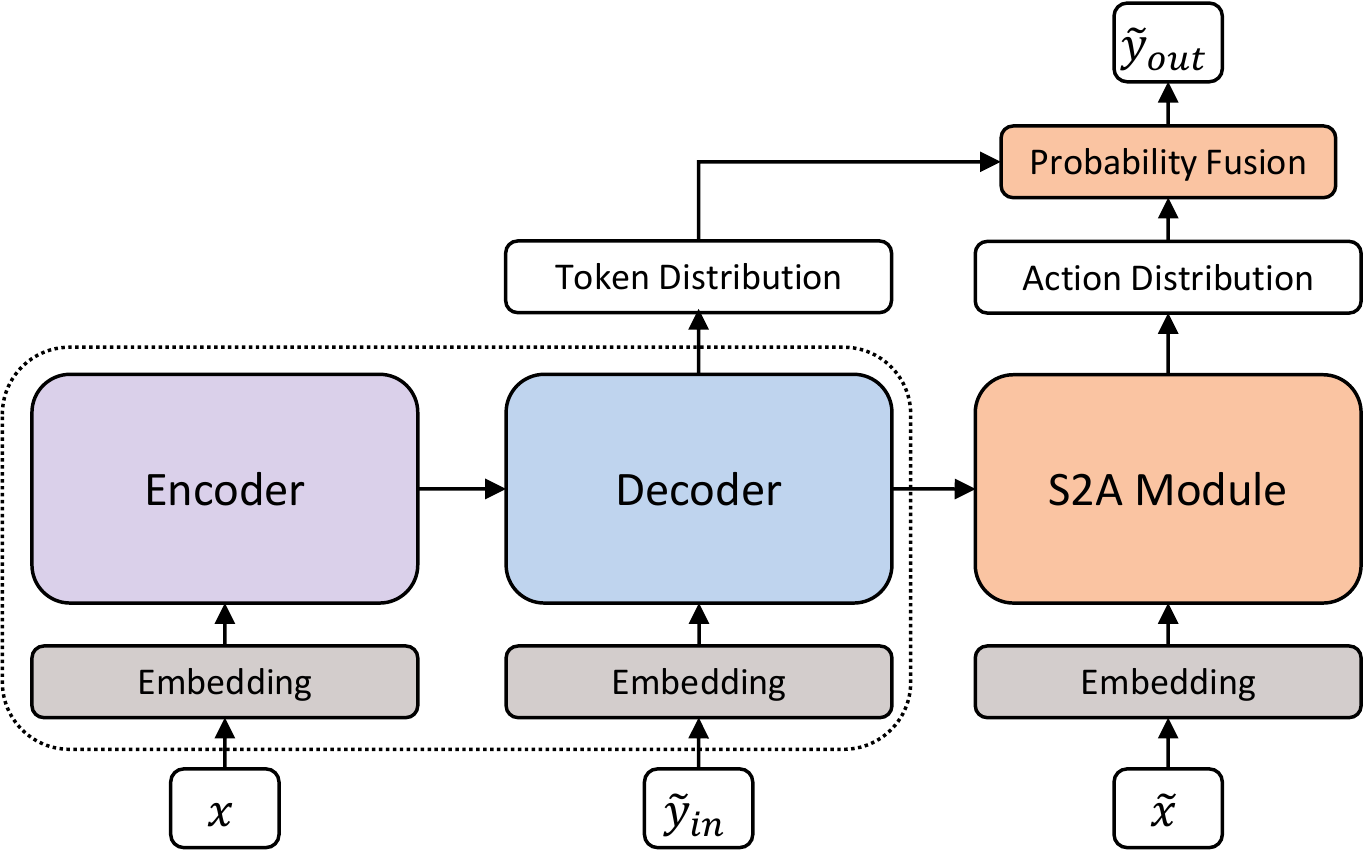}
    \caption{An illustration of the proposed framework.}
    \label{fig:model_pic}
\vspace{-10pt}
\end{figure}

We aim to build a framework that is able to alleviate the problems of both seq2seq models and sequence tagging models as discussed in the previous section. 
To be specific, the model should keep the original correct tokens as much as possible, while being able to dynamically generate editions w.r.t erroneous tokens without human knowledge.
To achieve these goals, we propose a novel Sequence-to-Action~(S2A) module based on the seq2seq framework, with tailored input formats.
An illustration of the proposed framework is shown in Figure~\ref{fig:model_pic}.

Specifically, our framework consists of three components: the encoder, the decoder and the S2A module. 
The S2A module is built upon the decoder to predict an action at each input position.
Different from previous sequence tagging works which usually introduce lots of labels and generate new tokens from the pre-defined vocabulary,
we simply design three actions named \texttt{SKIP}, \texttt{COPY} and \texttt{GEN}erate,
where \texttt{SKIP} indicates that the current input token is an error that should not appear in the output, \texttt{COPY} refers to an original correct token that should be kept, and \texttt{GEN} indicates a new token should be generated at this position, and the token is predicted from the whole dictionary instead of a pre-defined subset.

The S2A module takes the hidden output of the decoder and the source sentence as input, and we design a special input format to integrate them, ensuring that at each step, the action is taken considering both the source and target information. 
Then the action probability produced by the S2A module is fused with the traditional seq2seq prediction probability to obtain the final results. We introduce the details as follows.

\begin{table*}
\small
  \centering
  \begin{tabular}{l|llllllllll}
    \toprule
    
     & Sequence \\
    \midrule
    Source $x$ & \texttt{The} & \texttt{cat} & \texttt{\underline{is}} & \texttt{sat} & \texttt{\underline{at}} & \texttt{mat} & \texttt{.} & \texttt{[/S]} & & \\
    
    Gold $y$ & \texttt{The} & \texttt{cat} & \texttt{sat} & \texttt{\textbf{on}} & \texttt{\textbf{the}} & \texttt{mat} & \texttt{.} & \texttt{[/S]} & & \\
    
    Integrated $z$ & \texttt{The} & \texttt{cat} & \texttt{\underline{is}} & \texttt{sat} & \texttt{\underline{at}} & \texttt{\textbf{on}} & \texttt{\textbf{the}} & \texttt{mat} & \texttt{.} & \texttt{[/S]} \\
    
    Action $a$ & \texttt{C} & \texttt{C} & \underline{\texttt{S}} & \texttt{C} & \underline{\texttt{S}} & \textbf{\texttt{G}} & \textbf{\texttt{G}} & \texttt{C} & \texttt{C} & \texttt{C} \\
    
    $\tilde{x}$ & \texttt{The} &  \texttt{cat} &  \texttt{\underline{is}} &  \texttt{sat} & \texttt{\underline{at}}  & \texttt{mat} & \texttt{mat} &  \texttt{mat} & \texttt{.} &  \texttt{[/S]} \\
    
    $\tilde{y}_{\textrm{in}}$ & \texttt{[S]} & \texttt{The} & \texttt{cat} & \texttt{cat} & \texttt{sat} & \texttt{sat} & \texttt{\textbf{on}} &  \texttt{\textbf{the}} & \texttt{mat} & \texttt{.} \\
    
    $\tilde{y}_{\textrm{out}}$ & \texttt{The} & \texttt{cat} & \underline{\texttt{[BLK]}} & \texttt{sat} & \underline{\texttt{[BLK]}} & \texttt{\textbf{on}} &  \texttt{\textbf{the}} & \texttt{mat} & \texttt{.}  & \texttt{[/S]} \\
    \bottomrule
  \end{tabular}
\vspace{-10pt}
\caption{An illustration for the decoder input construction. We refer to \texttt{SKIP} as \texttt{S}, \texttt{COPY} as \texttt{C}, and \texttt{GEN} as \texttt{G}. 
To differentiate the three actions, we mark \texttt{SKIP} and \texttt{[BLK]} with underlines and \texttt{GEN} in boldface.
\texttt{[S]} and \texttt{[/S]} represent the begin- and end-of-sequence symbol.}
\label{tab:illustration}
\end{table*}

\subsection{Input Construction}

The encoder in our model is akin to a traditional seq2seq encoder, which takes $x$ as input and provides its hidden representation $h_e$.
As for the decoder side, instead of simply taking $y$ as input, 
we integrate $x$ with $y$ as the decoder input to provide direct information of the original sentence. 

Specifically, we follow four principles to integrate $x$ and $y$ into a new sequence $z$
with an action sequence $a$. 
For every token in $x$ and $y$, 
1) if the source token is an 
originally correct token (i.e., appears both in $x$ and $y$), we simply copy it to $z$ and assign the action \texttt{COPY} to it;
2) if the source token is an erroneous token that should be replaced (i.e., the corresponding correct token appears in $y$), we jointly append this token as well as the correct token to $z$, and assign \texttt{SKIP} to the erroneous token while assigning
\texttt{GEN} to the correct token;
3) if the source token is an erroneous token that should be deleted (i.e., no corresponding correct token appears in $y$), we append it to $z$ and assign the action \texttt{SKIP} to it;
4) for the target tokens that do not have alignments in $x$ and thus should be generated, we append them to $z$ and assign the action \texttt{GEN}.
Finally, we obtain the integrated sequence $z$ and the action sequence $a$, both with length $t$ and $t \geq \max (m,n)$.
This process can be implemented by dynamic programming and we provide an illustration in Table~\ref{tab:illustration}.

We then construct $\tilde{y}_{\textrm{in}}$ which serves as the decoder input and $\tilde{x}$ which maintains the source tokens and serves as the input to the S2A module.
Details are described in Algorithm~\ref{algo:train_seq_gen}. 
To construct $\tilde{x}$, we iterate over $z$ and keep the tokens with \texttt{COPY} and \texttt{SKIP} actions. For tokens with the \texttt{GEN} action, we fill up the location with its next token in the original source sentence. 
Meanwhile, for $\tilde{y}$, we keep the tokens with \texttt{COPY} and \texttt{GEN} actions, and fill up the locations with the \texttt{SKIP} action by introducing a special blank token \texttt{[BLK]}, which serves as a placeholder to represent the positions where the original tokens are skipped. 
It is worth noting that while training, we take $\tilde{y}$ as the target sequence $\tilde{y}_{\textrm{out}}$,
and we replace the \texttt{[BLK]} token with the most recent non-blank token on its left and take the right-shifted version of $\tilde{y}$ as the decoder input $\tilde{y}_{\textrm{in}}$ to enable teacher forcing.

\begin{algorithm}[tb]
\caption{Input Construction}
\label{algo:train_seq_gen}
\textbf{Input}: The source sentence $x$, the integrated sequence $z$ and the action sequence $a$ with length $t$ \\
\textbf{Output}: The S2A input $\tilde{x}$, the decoder input $\tilde{y}_{\textrm{in}}$, the target sequence $\tilde{y}_{\textrm{out}}$ \\
\begin{algorithmic}[1] 
\STATE {Initialize $\tilde{x}$ and $\tilde{y}$ as empty lists; $i=1$;}
\FOR{ $k\leftarrow 1$ \TO $t$}
    \IF{$a_k==\textrm{COPY}$}
        \STATE {Append $z_k$ to $\tilde{x}$; $i$++;}
        \STATE {Append $z_k$ to $\tilde{y}$;}
    \ELSIF{$a_k==\textrm{SKIP}$}
        \STATE {Append $z_k$ to $\tilde{x}$; $i$++;}
        \STATE {Append \textrm{BLK} to $\tilde{y}$;}
    \ELSE
        \STATE {Append $x_i$ to $\tilde{x}$;}
        \STATE {Append $z_k$ to $\tilde{y}$;}
    \ENDIF
\ENDFOR
\STATE {$\tilde{y}_{\textrm{out}} \leftarrow \tilde{y}$;}
\STATE {Replace \texttt{BLK} in $\tilde{y}$ with the token on its left;}
\STATE {$\tilde{y}_{\textrm{in}} \leftarrow $ \textrm{[S]} $ + \tilde{y}_{0:-1}$;}
\STATE \textbf{return} {$\tilde{x}$, $\tilde{y}_{\textrm{in}}$, $\tilde{y}_{\textrm{out}}$}
\end{algorithmic}
\end{algorithm}

\subsection{Training and Inference}

Given the source input $x$, the decoder input $\tilde{y}_{\textrm{in}}$, the S2A input $\tilde{x}$ and the target sentence $\tilde{y}_{\textrm{out}}$,
the computation flow of our model can be written as follows,
\begin{equation}
\label{equ:model_compute}
\begin{split}
    h = f_a(h_d, e(\tilde{x})) \quad &\text{where} \quad h_d = f_d(\tilde{y}_{\textrm{in}}, h_e) \\
    &\text{and} \quad h_e = f_e(x),
\end{split}
\end{equation}
where $f_a$, $f_d$ and $f_e$ indicate the S2A module, the decoder and encoder respectively, with the hidden outputs $h$, $h_d$ and $h_e$ correspondingly, and $e(\cdot)$ indicates the embedding lookup. 
Then the token prediction probability can be written as $p_d(\tilde{y}_{\textrm{out}}|\tilde{y}_{\textrm{in}}, x) = \textrm{softmax}(f_o(h_d))$, where $f_o(\cdot) \in \mathbb{R}^{d \times V}$ is the linear output layer that provides the prediction probability over the dictionary, while $d$ and $V$ indicate the dimension
of the hidden outputs and the vocabulary respectively.

The S2A module simply consists of two feed-forward layers, 
taking the concatenation of $h_d$ and $e(\tilde{x})$ as input and producing the probability over actions as output:
\begin{equation}
\label{equ:action_compute}
\begin{split}
   &p_a(a|h_d, e(\tilde{x})) \\
   &= \textrm{softmax}(\sigma([h_d; e(\tilde{x})] \cdot w_1 + b_1) \cdot w_2 + b_2),\\
   &\text{ where }  w_1 \in \mathbb{R}^{2d \times 2d}, w_2 \in \mathbb{R}^{2d \times 3}.
\end{split}
\end{equation}
For each position, the three actions \texttt{SKIP}, \texttt{COPY} and \texttt{GEN} indicate that the model should predict a \texttt{[BLK]} token, or predict the input token, or generate a new token respectively. Therefore, for the $i$-th position, denote the probabilities of three actions as $(p^i_a(s), p^i_a(c), p^i_a(g))=p^i_a$, we fuse them with the token probability $p^i_d$ to obtain the prediction probability of the S2A module $p_{\textrm{s2a}}^i \in \mathbb{R}^{V}$,
\begin{equation}
\begin{split}
    p&_{\textrm{s2a}}^i(\texttt{[BLK]}) = p^i_a(s), \\
    p&_{\textrm{s2a}}^i(\tilde{x}^i) = p^i_a(c), \\
    p&_{\textrm{s2a}}^i(V \setminus \{\texttt{[BLK]}, \tilde{x}^i\}) =  p^i_a(g) \cdot p^i_d(V \setminus \{\texttt{[BLK]}, \tilde{x}^i\}),
\end{split}
\end{equation}
where $\tilde{x}^i$ indicates the current input token
and $p^i_d(V \setminus \{\texttt{[BLK]}, \tilde{x}^i\})$ indicates the normalized token probability after setting the prediction of the blank token as well as the current token to $0$.
In this way, as the probabilities in $p_a$ ($3$-classes) are usually larger than that in $p_d$ ($V$-classes),
with action \texttt{COPY}, we amplify the probability of keeping the current token which is originally correct;
with action \texttt{GEN}, we force the model to predict a new token from the vocabulary;
otherwise with action \texttt{SKIP}, we force the model to skip the current erroneous token by predicting the \texttt{[BLK]} token.

Then the loss function of the proposed Sequence-to-Action module can be written as:
\begin{equation}
\label{equ:s2a-loss}
\begin{split}
    &L_{\textrm{s2a}}(\tilde{y}_{\textrm{out}}| \tilde{y}_{\textrm{in}}, \tilde{x}, x) \\
    &= -\sum_{i=1}^t \log p_{\textrm{s2a}}(\tilde{y}_{\textrm{out}}^i | \tilde{y}_{\textrm{in}}^{<i}, \tilde{x}^i, x; \theta_{\textrm{s2s}}, \theta_{\textrm{s2a}}),\\
\end{split}
\end{equation}
where $\theta_{\textrm{s2a}}$ indicates the parameters of the proposed S2A module. Together with the loss function of the traditional seq2seq model described in Equation~(\ref{equ:s2s-obj}), the final loss function of our framework can be written as:
\begin{equation}
\label{equ:total-loss}
\begin{split}
    L(x,y; \Theta) = &(1-\lambda) L_{\textrm{s2a}}(\tilde{y}_{\textrm{out}}| \tilde{y}_{\textrm{in}}, \tilde{x}, x; \theta_{\textrm{s2s}}, \theta_{\textrm{s2a}})) \\
    &+ \lambda L_{\textrm{s2s}}(\tilde{y}_{\textrm{out}} | \tilde{y}_{\textrm{in}}, x; \theta_{\textrm{s2s}}),
\end{split}
\end{equation}
where $\Theta=(\theta_{\textrm{s2s}}, \theta_{\textrm{s2a}})$ 
denotes the parameters of the model, and $\lambda$ is the hyper-parameter that controls the trade-off between the two 
objectives.

While inference, our model generates in a similar way as the traditional seq2seq model, except that we additionally take a source token 
$x^j$ 
as the input, which denotes the current source token that serves as the input to the S2A module. 
After the current generation step, if the predicted action lies in \texttt{SKIP} and \texttt{COPY}, we move the source token a step forward 
(i.e., $j$++). 
Otherwise with action \texttt{GEN}, we keep 
$j$
unchanged. In this way, we ensure that 
$x^j$
serves the same functionality as $\tilde{x}^i$ (i.e., the source token that should be considered at the current step) during training.

\begin{table*}
  \centering
  \begin{tabular}{l|ccc}
    \toprule
    Model & Precision & Recall & $F_{0.5}$\\
    \midrule
    Transformer Big & 64.9 & 26.6 & 50.4 \\
    LaserTagger~\citep{malmi2019encode} & 50.9 & 26.9 & 43.2 \\
    ESD+ESC~\citep{chen-etal-2020-improving-efficiency} & \textbf{66.0} & 24.7 & 49.5 \\
    \textbf{S2A Model} & 65.9 & \textbf{28.9} & \textbf{52.5}  \\
    Transformer Big (Ensemble) & 71.3 & 26.5 & 53.3  \\
    \textbf{S2A Model} (Ensemble) & \textbf{72.7} & \textbf{27.6} & \textbf{54.8}   \\
    \midrule
    Transformer Big + pre-train & 69.4 & 42.5 & 61.5 \\
    PRETLarge~\citep{kiyono-etal-2019-empirical} & 67.9 & 44.1 & 61.3 \\
    BERT-fuse~\citep{kaneko-etal-2020-encoder} & 69.2 & \textbf{45.6} & 62.6 \\
    Seq2Edits~\citep{stahlberg2020seq2edits} & 63.0 & \textbf{45.6} & 58.6 \\
    PIE~\citep{awasthi2019parallel} & 66.1 & 43.0 & 59.7 \\
    GECToR-BERT~\citep{omelianchuk2020gector} & 72.1 & 42.0 & 63.0 \\
    GECToR-XLNet~\citep{omelianchuk2020gector} & \textbf{77.5} & 40.1 & \textbf{65.3} \\
    ESD+ESC~\citep{chen-etal-2020-improving-efficiency} + pre-train & 72.6 & 37.2 & 61.0 \\
    \textbf{S2A Model} + pre-train & 74.0 & 38.9 & 62.7 \\
    Transformer Big + pre-train (Ensemble) & 74.1  & 37.5 & 62.0  \\
    GECToR (Ensemble)
     & \textbf{78.2} & \textbf{41.5} & \textbf{66.5}  \\
    \textbf{S2A Model} + pre-train (Ensemble) & 74.9 & 38.4 &  62.9  \\
    \bottomrule
  \end{tabular}
\caption{The results on the CoNLL-2014 English GEC task.
The top group of results are generated by models trained only on the BEA-2019 training set.
The bottom group of results are generated by models that are first pre-trained on a large amount of synthetic pseudo data followed with fine-tuning. Here we bold the best results of single models and ensemble models separately.
}
\label{tab:conll_res}
\end{table*}

\subsection{Discussion}
Our model combines the advantages of both the seq2seq and the sequence tagging models, while alleviating their problems.
On the one hand, while seq2seq models usually suffer from over-correction or omission of correct tokens in the source sentence~\citep{tu-etal-2016-modeling}, the proposed sequence-to-action module guarantees that the model will directly visit and consider the source tokens in $\tilde{x}$ as the next predictions' candidates
before making the final predictions.
On the other hand, comparing with sequence tagging models which depend on human-designed labels, lexical rules and vocabularies for generating new tokens, we only introduce three atomic actions in the model without other constraints, which enhances the generality and diversity of the generated targets. 
As shown in 
the case studies, sequence tagging models usually fail when dealing with hard editions, e.g., reordering or generating a long range of tokens, and our model performs well on these cases by generating the target sentence from scratch auto-regressively.
In addition, we choose the term ``action'' instead of ``label'' because the proposed sequence-to-action module does not need any actual labels to be trained, instead it is trained end-to-end by fusing the probability with the seq2seq model.

\section{Experiments}

\subsection{Datasets and Evaluation Metrics}
\label{dataset}

We conduct experiments on both Chinese GEC and English GEC tasks.
For the Chinese GEC task, we use the dataset of NLPCC-2018 Task 2~\citep{10.1007/978-3-319-99501-4_41}\footnote{\url{http://tcci.ccf.org.cn/conference/2018/taskdata.php}}, which is the first and latest benchmark dataset for Chinese GEC. 
Following the pre-processing settings~in \citep{DBLP:conf/aaai/ZhaoW20}, we get $1.2$M sentence pairs in all. Then $5$k sentence pairs are randomly sampled from the whole parallel corpus as the development set. The rest are used as the final training set.
We use the official test set\footnote{\url{https://github.com/pkucoli/NLPCC2018_GEC}} which contains $2$k sentences extracted from the PKU Chinese Learner Corpus. 
The test set also includes the annotations that mark
the golden edits of grammatical errors in each sentence.
We tokenize the sentence pairs following~\citep{DBLP:conf/aaai/ZhaoW20}. Specifically, we use the tokenization script of BERT\footnote{\url{https://github.com/google-research/bert}} to tokenize the Chinese symbols and keep the non-Chinese symbols unchanged. 

For English GEC, we take the datasets provided in the restricted track of the BEA-2019 GEC shared task \cite{bryant-etal-2019-bea}.
Specifically, the training set is the concatenation of the Lang-8 corpus~\citep{mizumoto-etal-2011-mining}, the FCE training set~\citep{yannakoudakis-etal-2011-new}, NUCLE~\citep{dahlmeier-etal-2013-building}, and W\&I+LOCNESS~\citep{granger2014computer,bryant-etal-2019-bea}.
We use the CoNLL-2013~\citep{ng-etal-2013-conll} test set as the development set to choose the best-performing checkpoint, which is then evaluated on the benchmark test set CoNLL-2014~\citep{ng-etal-2014-conll}.
We also conduct experiments 
by pre-training the model with $100$M synthetic parallel examples provided by~\citep{grundkiewicz-etal-2019-neural}.
All English sentences are preprocessed and tokenized by 32K SentencePiece \citep{kudo-richardson-2018-sentencepiece}
Byte Pair Encoding (BPE)~\citep{sennrich-etal-2016-neural}.

We use the official MaxMatch ($M^2$) scorer~\citep{dahlmeier-ng-2012-better}
to evaluate 
the models in both the Chinese and English GEC tasks. 
Given a source sentence and a system hypothesis, $M^2$ searches for a sequence of phrase-level edits between them that achieves the highest overlap with the gold-standard annotation. 
This optimal edit sequence is then used to calculate the values of precision, recall and $F_{0.5}$.

\begin{table*}
  \centering
  \begin{tabular}{l|ccc}
    \toprule
    Model & Precision & Recall & $F_{0.5}$\\
    \midrule
    YouDao~\citep{fu2018youdao} & 35.24 & \textbf{18.64} & 29.91 \\
    AliGM~\citep{zhou2018chinese} & 41.00 & 13.75 & 29.36 \\
    BLCU~\citep{ren2018sequence} & \textbf{41.73} & 13.08 & 29.02 \\
    Transformer & 36.57 & 14.27 & 27.86 \\
    \textbf{S2A Model} & 36.57 & 18.25 & \textbf{30.46} \\
    \midrule
    Transformer + MaskGEC~\citep{DBLP:conf/aaai/ZhaoW20} & \textbf{44.36} & 22.18 & 36.97 \\
    GECToR-BERT~\citep{omelianchuk2020gector} + 
    MaskGEC
    & 41.43 & 23.60 & 35.99 \\
    \textbf{S2A Model} + MaskGEC
    & 42.34 & \textbf{27.11} & \textbf{38.06} \\
    \bottomrule
  \end{tabular}
\caption{The results on the NLPCC-2018 Chinese GEC task.
The upper group of results are generated by models trained on the original NLPCC-2018 training data without data augmentation. The lower group of results are generated by models trained on the same training data but with the dynamic masking based data augmentation method proposed by MaskGEC~\citep{DBLP:conf/aaai/ZhaoW20}. Here we bold the best results.}
\label{tab:nlpcc_res}
\end{table*}

\subsection{Model Configurations}
\label{model}
For the basic seq2seq baseline, we adopt the original Transformer architecture. 
To compare with the previous works, we use the base/big model of Transformer for the Chinese/English GEC task respectively, and we follow
the official settings of Transformer: in the base/big model, the number of self-attention heads of Transformer is set to $8/16$, the embedding dimension is $512$/$1024$, the inner-layer dimension in the feed-forward layers is set to $2048$/$4096$ respectively.
For the loss function, we use the cross entropy loss with label smoothing and set the epsilon value to $0.1$.
We apply dropout~\citep{JMLR:v15:srivastava14a} on the encoders and decoders with probability of $0.3$. 
The implementations of our model and baselines are based on \texttt{fairseq}~\citep{ott2019fairseq} and will be available when published.

Moreover,  as suggested in~\citep{junczys-dowmunt-etal-2018-approaching},  for the English GEC task, we not only  report the results generated by single models, but also compare to 
ensembles of four models with different initializations.

\subsection{Baselines}
\label{baseline}

For the English GEC task, we compare the proposed S2A model to several representative systems, including three seq2seq baselines (Transformer Big, BERT-fuse~\citep{kaneko-etal-2020-encoder}, PRETLarge~\citep{kiyono-etal-2019-empirical}), four sequence tagging models (LaserTagger~\citep{malmi2019encode}, PIE~\citep{awasthi2019parallel}, GECToR~\citep{omelianchuk2020gector}, Seq2Edits~\citep{stahlberg2020seq2edits}), and a pipeline model ESD+ESC~\citep{chen-etal-2020-improving-efficiency}.
Specifically, for GECToR, we report their results when utilizing the pre-trained BERT model, XLNet model~\citep{yang2019xlnet} and  the results 
that integrate three different pre-trained language models in an ensemble. We denote them as GECToR-BERT,  GECToR-XLNet and GECToR (Ensemble) respectively.

For the Chinese GEC task, we compare S2A to several best performing systems evaluated on the NLPCC-2018 dataset, including three top systems in the NLPCC-2018 challenge (YouDao~\citep{fu2018youdao}, AliGM~\citep{zhou2018chinese}, BLCU~\citep{ren2018sequence}), the seq2seq baseline Char Transformer, and the current state-of-the-art method MaskGEC~\citep{DBLP:conf/aaai/ZhaoW20}.
Note that the proposed S2A model is orthogonal to MaskGEC, and we also report our results enhanced with the data augmentation method of MaskGEC.

In addition, we reproduce and conduct Chinese GEC experiments with the sequence tagging based method GECToR~\citep{omelianchuk2020gector}, which is originally designed for the English task.
To adapt it to the Chinese GEC task, we utilize the pre-trained Chinese BERT model \texttt{BERT-base-Chinese} and use \$KEEP, \$REPLACE and \$APPEND as the tags.
We pre-train the model using the data augmentation method proposed in MaskGEC and then fine-tune it on the training set. 
We follow the default hyper-parameter settings, and we set the max iteration number to $10$ while inference.

\subsection{Results}
\label{results}

The English GEC results are summarized in Table~\ref{tab:conll_res}. In the top group of results without pre-training, the proposed model achieves
$52.5 / 54.8$ in $F_{0.5}$ score with single/ensemble model, which significantly outperforms the baselines.
Compared to Transformer Big, 
our model achieves an improvement of $2.1$ in $F_{0.5}$ score with single model, and an improvement of $1.5$ in $F_{0.5}$ score with ensemble model.
Further, when pre-trained on the synthetic data, our method still achieves clear gains over most of the listed models. In the meantime, it slightly outperforms BERT-fuse using the pre-trained BERT in their framework, which is not used in the S2A model. One can also notice that, as benefits of pre-trained language models, GECToR-XLNet performs the best among all the listed 
methods with single models, and GECToR (Ensemble) performs the best among all the baselines. Nevertheless, the proposed S2A model uses neither any pre-trained language models, nor human designed rules, and is only outperformed by the GECToR models.

We proceed with the Chinese GEC results as shown in Table~\ref{tab:nlpcc_res}. 
In the upper group, when trained on the NLPCC-2018 training set without data augmentation, our proposed model consistently outperforms all the other baseline systems. In the lower group, when trained with the data augmentation method, our model outperforms MaskGEC with an improvement of $1.09$ in $F_{0.5}$ score and achieves a new state-of-the-art result.

It is worth noting that GECToR, 
which performs the best on the English GEC task, degenerates when generalizing to the Chinese GEC task.
Without a well-designed edit vocabulary in Chinese GEC, it fails to achieve comparable performance as a standard seq2seq model even equipped with a pre-trained Chinese BERT model. In comparison, the proposed S2A framework is language independent with good generality.

\begin{table}[htb]
  \centering
  \begin{tabular}{l|ccc}
    \toprule
    Model & \texttt{COPY} & \texttt{SKIP} & \texttt{GEN}\\
    \midrule
    Transformer & 96.6 & 37.2 & 13.9 \\
    \midrule
    GECToR & \textbf{96.8} & \textbf{43.3} & 12.1 \\
    \midrule
    \textbf{S2A Model} & 96.7 & 39.8 & \textbf{14.7} \\
    \bottomrule
  \end{tabular}
\caption{$F_1$ values of \texttt{COPY}, \texttt{SKIP} actions and \texttt{GEN} action fragments generated by Transformer, GECToR, S2A model on the NLPCC-2018 test set. }
\label{tab:ananlysis}
\end{table}

\begin{figure*}[htb]
    \centering
    \includegraphics{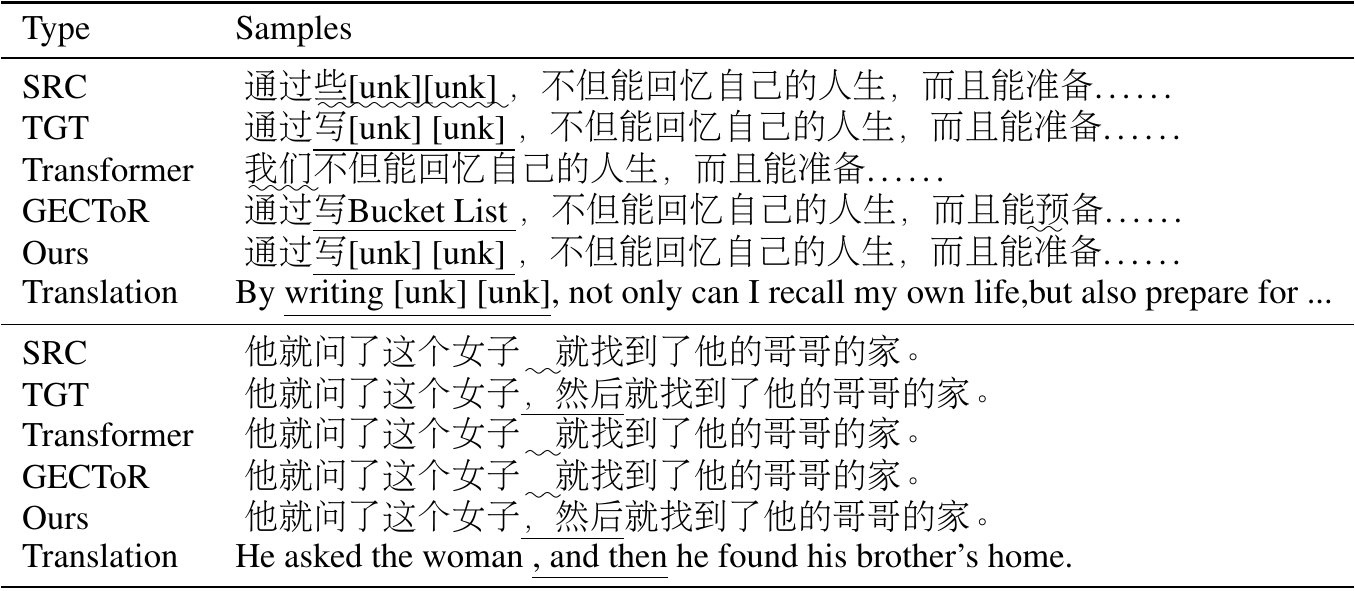}
\caption{
Case studies of the seq2seq (Transformer) model, the sequence tagging model (GECToR) and the proposed S2A model on the Chinese NLPCC-2018 test set. The translation of the golden target in each pair is also listed. 
The tokens with wavy lines are errors, while tokens with underlines are the corrections made by the gold target or hypotheses.
\texttt{[unk]} means out-of-vocabulary words.}
\label{fig:case_study}
\end{figure*}

\subsection{Accuracy of Action Prediction}
\label{analysis}

In order to analyze the impact of the S2A module, we evaluate the generated actions. 
To compare with other models, we extract the action sequences from the results generated by them on the NLPCC-2018 test set.
Next, we calculate the 
$F_1$ values of \texttt{COPY}, \texttt{SKIP} actions and \texttt{GEN} action fragments.
All scores are listed in Table~\ref{tab:ananlysis}. 

As shown in Table~\ref{tab:ananlysis}, with a lower $F_{0.5}$ value 
of the $M^2$ score, GECToR performs the best in deciding whether each token should be copied or skipped from the original sentence. This is not surprising considering that GECToR is a sequence editing model. In the meantime, with an extra S2A module, 
our proposed framework can learn to explicitly process each token from the original sentence as GECToR does, and it performs better than Transformer in predicting \texttt{COPY} and \texttt{SKIP}. 
As for the action 
\texttt{GEN}, S2A and Transformer both generate new tokens in an auto-regressive way, which are more flexible than GECToR. As a consequence, they achieve higher accuracy when predicting the action \texttt{GEN} than GECToR, and the S2A model performs the best in this aspect.

If we look further into one of the gold-standard annotations in NLPCC 2018, among all the corrections, $1864$ are replacing error characters, $130$ are reordering characters, $1008$ are inserting missing characters, and $789$ are deleting redundant characters. Except the corrections of deleting redundant characters, the other three types of corrections all involve both deleting and extra generating. As a consequence, improving the accuracy of predicting \texttt{GEN} may help more to boost the GEC quality, which is the advantage of the proposed S2A model, and agrees with the results in Table~\ref{tab:nlpcc_res}.

\subsection{Case Study}
\label{case_study}

Next, we conduct case studies to intuitively demonstrate the advantages of the proposed S2A model. 
All the cases are picked from the NLPCC-2018 Chinese GEC test set. Results are listed in Figure~\ref{fig:case_study}.

\noindent \textbf{Comparison to Seq2seq Methods}~~~
In the upper part of Figure~\ref{fig:case_study}, we list
an example 
in which a standard seq2seq model does not perform well.
In this example, the seq2seq model suffers from word omissions. 
In comparison, the proposed S2A model rarely omits a long string of characters.

\noindent \textbf{Comparison to Sequence Tagging Methods}~~~
In the bottom example of Figure~\ref{fig:case_study}, a relatively long insertion is necessary which is very common in Chinese GEC.
As shown in the results, the sequence tagging model is not able to produce correct output in this challenging case, while the proposed S2A model can generate the expected result due to its flexibility, which justifies the generality of S2A.

\section{Conclusion}

In this paper, we propose a Sequence-to-Action (S2A) model based on the sequence-to-sequence framework for Grammatical Error Correction. 
We design tailored input formats so that in each prediction step, the model learns to \texttt{COPY} the next unvisited token in the original erroneous sentence, \texttt{SKIP} it, or \texttt{GEN}erate a new token before it. 
The S2A model alleviates the over-correction problem, and does not rely on any human-designed rules or vocabularies, which provides a language-independent and flexible GEC framework. 
The superior performance on both Chinese and English GEC justifies the effectiveness and generality of the S2A framework. 
In the future, we will extend our idea to other text editing tasks such as text infilling, re-writing and text style transfer.

\section{Acknowledgments}
This research was supported by Anhui Provincial Natural Science Foundation (2008085J31).

\bibliography{aaai22.bib}

\clearpage

\appendix

\section{Appendix}

\subsection{Dataset Statistics}

We list the statistics of all the datasets used in this paper in the following tables. Specifically, Table~\ref{tab:nlpcc} and Table~\ref{tab:bea} include the detailed information for the datasets used in the Chinese GEC task and the English GEC task respectively.

\begin{table}[H]
\centering
    \caption{Dataset in the NLPCC 2018 Task 2 }
    \label{tab:nlpcc}
    \centering
    \begin{tabular}{lll}
        \toprule
        Corpus & Split & \#Sent. \\
        \midrule
        Lang-8 & Train & 1.2M  \\
        Lang-8 & Valid & 5,000  \\
        PKU CLC & Test & 2,000 \\
        \bottomrule
    \end{tabular}
\end{table}

\begin{table}[H]
\centering
    \caption{Datasets in the BEA-2019 Shared Task}
    \label{tab:bea}
    \centering
    \begin{tabular}{lll}
    \toprule
    Corpus & Split & \#Sent.\\
    \midrule
    Lang-8 & Train & 1.1M \\
    NUCLE & Train & 57K \\
    FCE & Train & 32K \\
    W\&I+LOCNESS & Train & 34K \\
    
    CoNLL-2013 & Valid & 1.3K \\
    CoNLL-2014 & Test & 1.3K \\
    \bottomrule
    \end{tabular}
\end{table}

\subsection{Training Details}

Here we also report the training details in our experiments.
For both English/Chinese GEC tasks, we train the S2A model with 4 Nvidia V100 GPUs.
Models are optimized with the Adam optimizer~\citep{kingma2015adam} with a beta value of ($0.9$, $0.98$)/($0.9$, $0.998$).

For the English GEC Task, the max tokens per GPU is set to $4096$ with the update frequency set to $2$. We set the learning rate to $1e-4$ when the training is only conducted on datasets from the BEA-2019 shared task. And we train it for $30K$ updates. When synthetic data is used, we first pre-train our model on the synthetic data for $4$ epochs with the max tokens per GPU set to $3072$ and the update frequency set to 2. After that, we fine-tune it on the BEA-2019 training datasets with the same settings as when no synthetic data is introduced. The hyper-parameter $\lambda$ in the English GEC task is set to $0.4$. 

For the Chinese GEC Task, we use the polynomial learning rate decay scheme, with $8$k warm-up steps, whereas for the English GEC Task, we use the inverse sqrt learning rate decay scheme, with $4$k warm-up steps. The max tokens per GPU is set to $8192$. And we set the learning rate to $4e-4$. For the experiments without data augmentation, we train it for $80K$ steps, while 
the for experiments with the dynamic masking data augmentation technique, we train it for $120K$ steps. The hyper-parameter $\lambda$ in the Chinese GEC task is set to $0.05$.

\subsection{Influence of $\lambda$}
The $\lambda$ value mentioned above in Equation~(\ref{equ:total-loss}) 
is chosen on the development set.
To investigate the actual influence of the hyper-parameter $\lambda$ that controls the trade-off between the two objectives in Equation~(\ref{equ:total-loss}),
we conduct experiments on the English GEC task. We fix the rest of the hyper-parameters of the model and 
select the values of
$\lambda$ from $[0, 0.05, 0.1, 0.2, 0.4, 0.6]$.

\begin{figure}
    \centering
    \includegraphics[width=0.85\columnwidth]{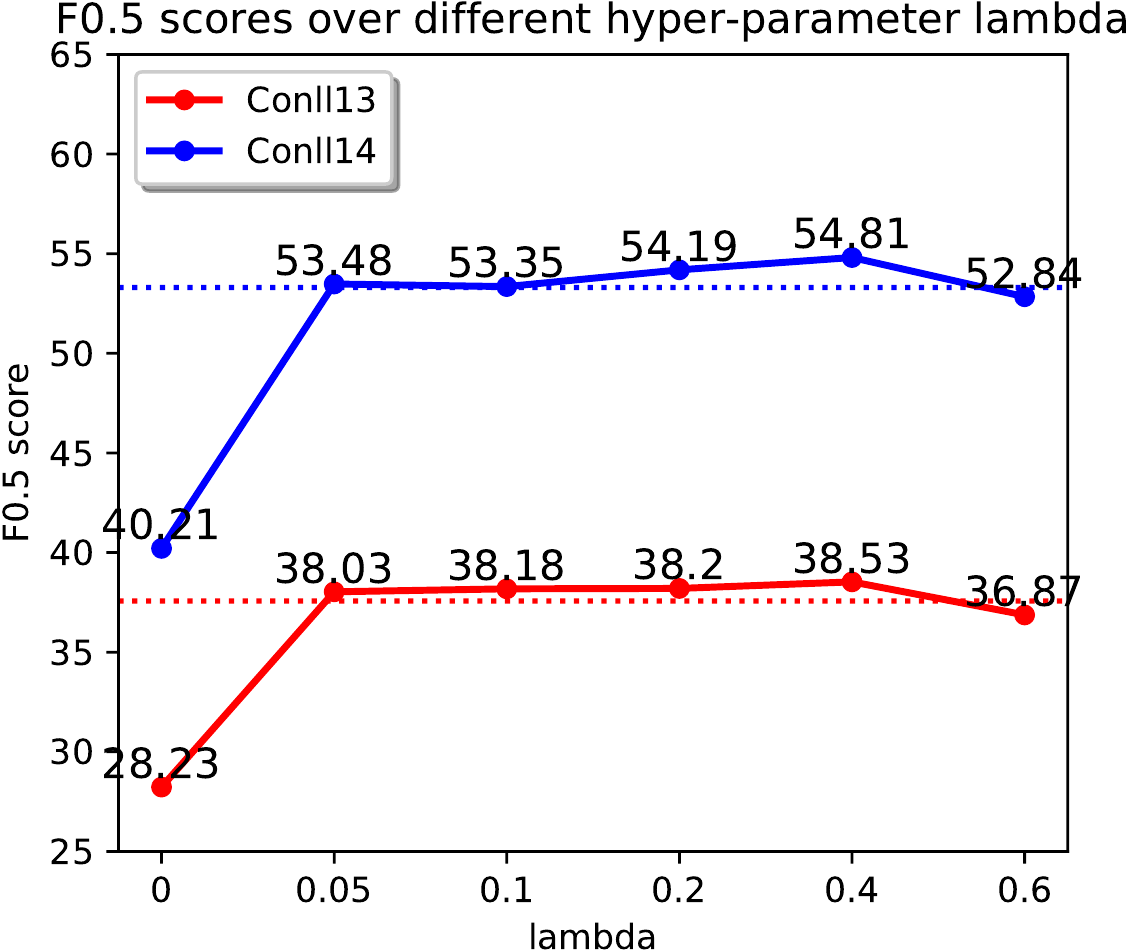}
\caption{
The $F_{0.5}$ scores on the development set (the red line) and the test set (the blue line) of the English GEC task over different $\lambda$ values.
The performance of the Transformer (Big) baseline is plotted as dashed lines.}
\label{fig:ablation}
\end{figure}

We plot the $F_{0.5}$ scores of the S2A model and Transformer (Big) baseline on the development and test set (i.e., CoNLL-2013~\citep{ng-etal-2013-conll} and CoNLL-2014~\citep{ng-etal-2014-conll}) in Figure~\ref{fig:ablation}.
As is shown in Figure~\ref{fig:ablation}, the proposed S2A model performs stably with varying nonzero $\lambda$ values and achieves the best results 
when $\lambda=0.4$.
We also find that when $\lambda=0$, which means that the seq2seq loss $\mathcal{L}_{\textrm{s2s}}$ is omitted, the performance of the S2A model decreases drastically, indicating the original seq2seq loss is crucial for predicting tokens.

\subsection{More Case Studies}

\begin{figure*}[htb]
    \centering
    \includegraphics{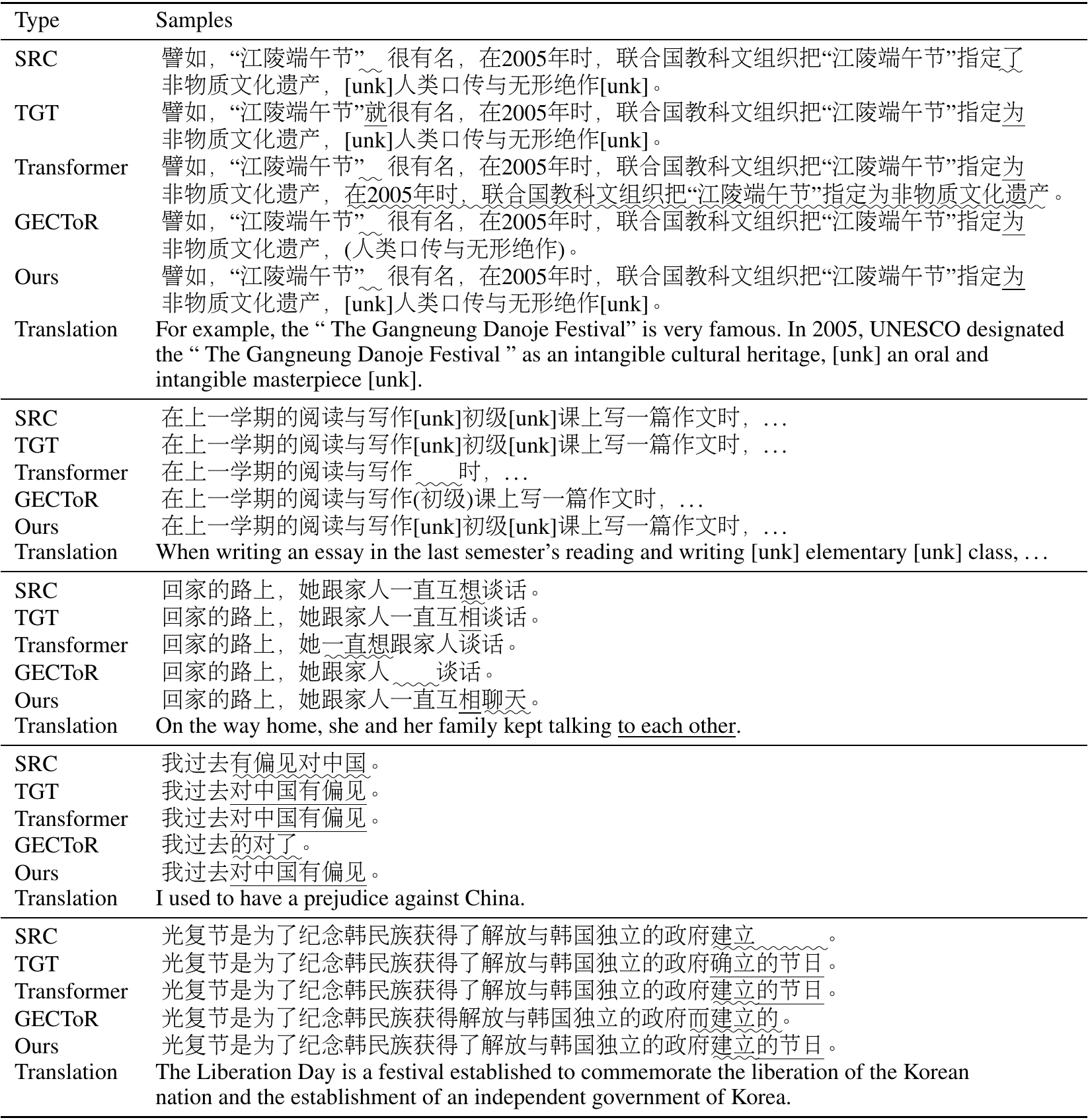}
\caption{
More case study of the seq2seq (Transformer) model, the sequence tagging model (GECToR) and the proposed S2A model on the Chinese NLPCC-2018 test set. The translation of the golden target in each pair is also listed. 
The tokens with wavy lines are errors, while tokens with underlines are the corrections made by the gold target or hypotheses.
\texttt{[unk]} means out-of-vocabulary words.}
\label{fig:more_case_study}
\end{figure*}

In addition to the case studies in the main paper, we list six more cases in Figure~\ref{fig:more_case_study}. 
In the first example, a long string of characters are generated twice in the seq2seq model, resulting a severe repetition problem. 
In the second example, the seq2seq model wrongly omits several originally correct words.
These situations nearly never happen in GECToR and our S2A model.

The third example corresponds to the cases where over-correction happens for the seq2seq model. 
In the third example, the correction produced by the seq2seq model means that ``she always wanted to talk ''. 
In contrast, thanks to the guidance from the proposed sequence-to-action module, our model tends to follow the original structure, thus alleviating such biases in generation.

For the fourth and fifth example, 
longer reordering or insertions are necessary. GECToR, which is the sequence tagging model, can not handle these situations essentially. In comparison, our S2A model can perform as well as a seq2seq model.

\end{document}